\documentclass[12pt]{article}
\usepackage{graphicx}
\usepackage{epstopdf}
\usepackage{amssymb}
\usepackage{amsmath,epsfig}
\usepackage[authoryear,round]{natbib} 
\newtheorem{assump}{Assumption}

\newtheorem{proposition}{Proposition}
\newtheorem{corollary}{Corollary}

\def\btheta{\mbox{\boldmath $\theta$}}

\def\bSigma{\mbox{\boldmath $\Sigma$}}
\def\bbeta{\mbox{\boldmath $\beta$}}

\def\mR{\mathbb{R}}

\def\bmu{\mbox{\boldmath $\mu$}}
\def\bepsilon{\mbox{\boldmath $\epsilon$}}
\def\bb{{\bf b}}
\def\bz{{\bf z}}

\def\bI{{\bf I}}

\def\bQ{{\bf Q}}
\def\bR{{\bf R}}
\def\bX{{\bf X}}
\def\bW{{\bf W}}

\def\bw{{\bf w}}

\def\bu{{\bf u}}
\def\bx{{\bf x}}
\def\bv{{\bf v}}
\def\by{{\bf y}}
\def\nn{\nonumber}

\def\v2{\vspace{0.2in}}

\begin{document}
\setcounter{page}{1}
\baselineskip=17pt
\footskip=.3in

\title{Large scale analysis of generalization error in learning using  margin based classification methods}

\author{Hanwen Huang \\\\
    {\it Department of Epidemiology and Biostatistics}\\
    {\it University of Georgia, Athens, GA 30602}\\
    huanghw@uga.edu\\\\
     Qinglong Yang\footnote{Corresponding author} \\\\
       {\it School of Statistics and Mathematics}\\
       {\it Zhongnan University of Economics and Law}\\
       {\it Wuhan, Hubei 430073, P. R. China}\\
      yangqinglong@zuel.edu.cn}
\date{}

\maketitle

\begin{abstract}
Large-margin classifiers are popular methods for classification. We derive the asymptotic expression for the generalization error of a family of large-margin classifiers in the limit of both sample size $n$ and dimension $p$  going to $\infty$ with fixed ratio $\alpha=n/p$. This family  covers a broad range of commonly used classifiers including support vector machine, distance weighted discrimination, and penalized logistic regression. Our result can be used to establish the phase transition boundary for the separability of two classes.  We assume that the data are generated from a single multivariate Gaussian distribution with arbitrary covariance structure. We explore two special choices for the covariance matrix: spiked population model and two layer neural networks with random first layer weights. The method we used for deriving the closed-form expression is from statistical physics known as the replica method. Our asymptotic results match simulations already when $n,p$ are of the order of a few hundreds. For two layer neural networks, we reproduce the recently developed `double descent' phenomenology for several classification models. We also discuss some statistical insights that can be drawn from these analysis.
\end{abstract}
 
\noindent%
\vfill

\section{Introduction}

Classification is a very useful supervised learning technique for information extraction from data. The goal of classification is to construct a classification rule based on a training set where both covariates and class labels are given. Once obtained, the classification rule can then be used for class prediction of new objects whose covariates are available. There are a large number of methods for classification in the literature. Examples include Fisher linear discrimination analysis, logistic regression, k-nearest neighbor, decision trees, neural networks, boosting, and many others. See \cite{hastie01statisticallearning} for more comprehensive reviews of various classification methods. Among numerous classification techniques, margin-based classifiers have attracted tremendous attentions in recent years due to their competitive performance and ability in handling high dimensional data. The margin-based classifiers focus on the decision boundaries and bypass the requirement of estimating the class probability given input for discrimination.

The support vector machine (SVM) is one of the most well known large margin classifiers. Since its introduction, the SVM has gained much popularity in both machine learning and statistics. However, as pointed out by \cite{Marron2007}, SVM may suffer from a loss of generalization ability in the high-dimension-low-sample size (HDLSS) setting due to data-piling problem. They proposed distance weighted discrimination (DWD) as a superior alternative to SVM. \cite{liu:sigclust} proposed a family of large-margin classifiers, namely, the large-margin unified machine (LUM) which embraces both SVM and DWD as special cases. Besides SVM, DWD, and LUM, there are a number of other large margin classifiers introduced in the literature. Examples include the penalized logistic regression (PLR) \citep{wahba1999,lin2000}, $\psi$-learning \citep{shen2003}, the robust SVM \citep{yichao}, and so on. 

Despite some known properties of these methods, a practitioner often needs to face one natural question: which method should one choose to solve the classification problem in hand?  The choice can be difficult because typically the behaviors of different classifiers vary from setting to setting. Most of the previous studies in this area are empirical. For example, simulation and real data analysis indicate that DWD performs better than SVM especially in HDLSS cases, see e.g. \cite{Benito04,Qiao2011,qiao15,zou2,zou1}. Also simulation studies in \cite{liu:sigclust} have shown that soft classifiers tend to give more accurate classification results when the true probability functions are relatively smooth. Despite such substantial effort, not too much theoretical studies have been conducted to quantitatively characterize the performance of different classification methods. 

The objective of this paper is  to follow up on a recent wave of research works aiming at providing sharp performance characterization of classical statistical learning methods including regression, classification, and principle component analysis. Particularly, we derive the asymptotic behavior of margin based classification methods in the limit of both large sample size $n$ and large dimension $p$ with fixed ratio $\alpha=p/n$. The main literature related to this work is represented by a series recent papers which derive asymptotic results for classification  in the joint limit $p,n\rightarrow\infty$ with $n/p=\alpha$. \cite{Huang17,maistatistical} studied SVM under Gaussian mixture models in which the data are assumed to be generated from Gaussian mixture distribution with two components, one for each class. The covariance matrix is assumed to follow a spiked population model. Under the same setting,  \cite{mailiao,huang2019large} studied regularized logistic regression and general margin based classification methods respectively. \cite{montanari2019generalization} studied the hard margin SVM under the single Gaussian model in which the data are assumed to be generated from a single Gaussian distribution. \cite{goldt2019modelling} studied the regularized logistic regression under the single Gaussian model with covariance structure generated from two layer neural network model with random first layer weights.

In this paper, we derive the asymptotic performance of general margin based classification method under the single Gaussian model with arbitrary covariance structure. Our result is quite general in the sense that the family covers many of the aforementioned classifiers such as SVM, DWD, and PLR. Moreover, the covariance structure also includes spiked population model and two layer neural network model as special cases. We derive the analytical results using the replica method developed in statistical mechanics. Our result provides some insights on the behavior change among different classification methods. It also helps to shed some light on how to select the best model and optimal tuning parameter for a given classification task. As a corollary, we derive the phase transition boundary for the separability of two classes which embraces the previous results in \cite{candes2020} and \cite{9022461} as special cases. 

Moreover, for the two layer neural network covariance structure, our results exhibit the recently developed `double descent' phenomenon which has been demonstrated
empirically in \cite{Belkin15849}. It is referred to as a peculiar behavior of the test error as a function of  overparametrization ratio $\psi_1=p/n$. Namely, the test error peaks at a critical value of $\psi_1$ where the training error vanishes, and descends again after that. This picture have been theoretically studied in \cite{belkin2019models,double3,double4} for simple least square estimators. It was also studied in \cite{mei2019generalization} for nonlinear regression and in \cite{goldt2020gaussian} for logistic regression. Here we can reproduce this phenomenon for general margin based classification methods.

The rest of this paper is organized as follows. In Section \ref{main}, we first present the general result for the asymptotic generalization error of margin based classification methods and then apply it to two special covariance structures: spiked population model and two layer neural network model. The phase transition boundaries under different settings for the separability of two classes are also discussed. In Section \ref{numeric}, we demonstrate the numerical analysis of prediction error and compare them with the simulation results based on finite size system. Some discussion is provided in Section \ref{discussion}. The technical proofs are collected in the appendix.

\section{Main analytical results}\label{main}

\subsection{Overview of the Margin-Based Classification Method}\label{method}
In the binary classification problem, we are given a training dataset consisting of $n$ observations $\{(\bx_i,y_i); i=1,\cdots,n\}$ where $\bx_i\in \mR^p$ represents the input vector and $y_i\in\{+1,-1\}$ denotes the corresponding output class label, $n$ is the sample size, and $p$ is the dimension. Assume that the data are drawn i.i.d from an unknown joint probability distribution $P(\bx,y)$.  

The goal of linear classification is to find a linear function $f(\bx)=\bx^T\btheta$ with $\btheta\in\mR$ and predict the class labels using sign$(f(\bx))$. Define the functional margin as $yf(\bx)$ which is larger than 0 if correct classification occurs. In this paper, we focus on large-margin classification methods which can be fit in the regularization framework of Loss + Penalty. The loss function is used to keep the goodness of fit to the data while the penalty term is to avoid overfitting. Using the functional margin, the regularization formulation of binary large-margin classifiers can be summarized as the following optimization problem
\begin{eqnarray}\label{class}
\hat{\btheta}&=&\text{argmin}_{\btheta\in\mR^p}\left\{\sum_{i=1}^nV(y_i\bx^T_i\btheta)+\sum_{j=1}^pJ_\tau(\theta_j)\right\},
\end{eqnarray}
where $V(\cdot)\ge 0$ is a loss function, $J_\tau(\cdot)$ is the regularization term, and $\tau\textgreater 0$ is the tuning parameter for penalty. 

The general requirement for loss function is convex decreasing and $V(u)\rightarrow 0$ as $u\rightarrow\infty$. Many commonly used classification techniques can be fit into this regularization framework. The examples include penalized logistic regression (PLR; \cite{lin2000}), support vector machine (SVM; \cite{Vapnik95}), and distance weighted discrimination (DWD; \cite{Marron2007}). The loss functions of these classification methods are
\begin{eqnarray}\nn
\text{PLR}:&&V(u)=\log[1+\exp(-u)],\\\nn
\text{SVM}:&&V(u)=(1-u)_+,\\\nn
\text{DWD}:&&V(u)=\left\{\begin{array}{ccc}1-u&if&u\le \frac{1}{2}\\\frac{1}{4u}&if&u\textgreater\frac{1}{2}\end{array}\right..
\end{eqnarray}
Besides the above methods, many other classification techniques can also be fit into the regularization framework, for example, the large-margin unified machine \citep{liu2011}, the AdaBoost in Boosting \citep{FREUND1997119,friedman2000}, the import vector machine (IVM; \cite{doi:10.1198/106186005X25619}), and $\psi$-learning \citep{shen2003}. 

The commonly used penalty functions include $J_\tau(\theta)=\frac{\tau}{2}\theta^2$ for $L_2$ regularization and $J_\tau(\theta)=\tau|\theta|$ for sparse $L_1$ regularization. In this paper, we focus on the standard $L_2$ regularization.

\begin{figure}[hbtp]
    \begin{center}
      \epsfig{file=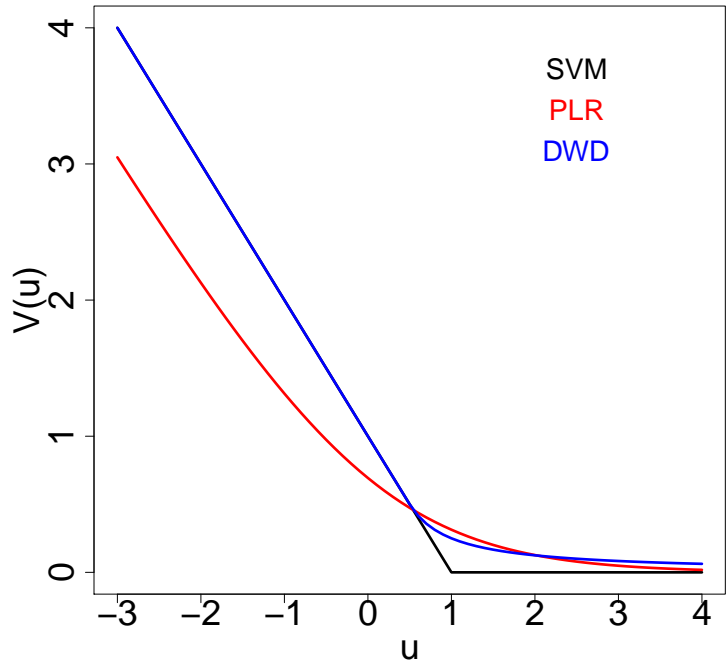,width=9.1cm,angle=-90}
    \end{center}
    \caption{Plots of various loss functions.}
    \label{figure0}
\end{figure}

Figure \ref{figure0} displays three loss functions: PLR, SVM, and DWD. Note that all loss functions have continuous first order derivatives except the hinge loss of SVM which is not differentiable at $u=1$. Among the three loss functions, PRL has all order derivatives while DWD only has first order derivative. As $u\rightarrow -\infty$, $V(u)\rightarrow -u$ for all methods.  As $u\rightarrow\infty$, $V(u)$ decays to 0 but with different speeds. The fastest one is SVM, followed by PLR and DWD. We will see in Section \ref{numeric} that the decay speed of the loss function has big influence on the classification performance in situations where the tuning parameter $\tau$ is small. 

\subsection{Asymptotic generalization error}\label{perf}
For the training data, denote the design matrix  as $\bX=[\bx_1,\cdots,\bx_n]^T$ and the response vector as $\by=[y_1,\cdots,y_n]$. Let the test error be defined by
\begin{eqnarray}\nn
{\cal E}(\by,\bX)=P(y_{new}\bx_{new}^T\hat{\btheta}(\by,\bX)\le 0),
\end{eqnarray}
where expectation is with respect to a fresh sample $(y_{new},\bx_{new})$ independent of the training data $(\by,\bX)$. We will sometimes refer to ${\cal E}(\by,\bX)$ as to the prediction error. We will determine the precise asymptotics of the test error in the limit of $n,p\rightarrow\infty$ with $n/p\rightarrow\alpha\in(0,\infty)$. 

We assume covariates $\bx_i\sim N(0,\bSigma)$ to be independent draws from a $p$-dimensional centered Gaussian with covariance $\bSigma$ and responses to be distributed according to 
\begin{eqnarray}\nn
P(y_1=+1|\bx_i)=1-P(y_1=-1|\bx_i)=g(\bx_i^T\btheta_\star)
\end{eqnarray}
for some vector $\btheta_\star\in\mR^p$ and monotone nonlinear function $g(\cdot)$: $\mR\rightarrow[0,1]$. In what follows we will index sequence of instances by $n\in N$, and it will be understood that $p=p_n$.
In order for the limit to exist and be well defined, we need to make specific assumptions about the behavior of the covariance matrix $\bSigma=\bSigma_n$ and the true parameters vector $\btheta_\star=\btheta_{\star,n}$. Let $\bSigma_n=\sum_{j=1}^p\lambda_j\bv_j\bv_j^T$ be the eigenvalue decomposition of $\bSigma$ with $\lambda_1\ge\lambda_2\ge\cdots\ge\lambda_p$ and $\bv_j\in\mR^p$ being orthonormal vectors for $1\le j\le p$. Similar to \cite{montanari2019generalization}, our first assumption requires that $\bSigma$ is well conditioned.
\begin{assump}\label{assum1}
Let $\lambda_{min}(\bSigma_n)=\lambda_p(\bSigma_n)$ and $\lambda_{max}(\bSigma_n)=\lambda_1(\bSigma_n)$, then $\lambda_1(\bSigma_n)=O_p(1)$ and $\lambda_p(\bSigma_n)=O_p(1)$.
\end{assump}
Assumption \ref{assum1} indicates that there exist constants $C_1,C_2\in(0,\infty)$ such that,  
\begin{eqnarray}\nn
C_1\le\lambda_{min}(\bSigma_n)\le\lambda_{max}(\bSigma_n)\le C_2.
\end{eqnarray}
Our second assumption concerns the eigenvalue distribution of $\bSigma_n$ as well as the decomposition of $\btheta_{\star,n}$ in the basis of eigenvectors of $\bSigma_n$. 
\begin{assump}\label{assum2}
Let $\lim_{n\rightarrow\infty}\|\btheta_{\star,n}\|_2=c$, $\rho_n=(\btheta_{\star,n}^T\bSigma_n\btheta_{\star,n})^{1/2}$, and $ w_j=\sqrt{p\lambda_j}\btheta_{\star,n}^T\bv_j/\rho_n$. Then the empirical distribution of $\{(\lambda_j, w_j)\}_{1\le j\le p}$ converges to a probability distribution $\mu$ on $\mR_{\textgreater 0}\times\mR$
\begin{eqnarray}\nn
\frac{1}{p}\sum_{j=1}^p\delta_{\lambda_j, w_j}\rightarrow\mu.
\end{eqnarray}
In particular, $\int w^2\mu(d\lambda,dw)=1$, and $\rho_n\rightarrow\rho$, where $1/\rho^2=\int(w^2/c\lambda)\mu(d\lambda,dw)$.
\end{assump}

Let us begin by introducing some functions.  For a given loss function $V(u)$, we define the proximal operator function
\begin{eqnarray}\label{conj}
\psi(a,b)=\text{argmin}_u\left\{V(u)+\frac{(u-a)^2}{2b}\right\},
\end{eqnarray}
for $b\textgreater 0$ which can be considered as the solution of equation 
\begin{eqnarray}\nn
\partial V(u)+\frac{u-a}{b}=0,
\end{eqnarray}
where $\partial V(u)$ is one of the sub-gradients of $V(u)$. For convex $V(u)$, this equation has unique solution. Specifically, for SVM loss, we have closed form expression
\begin{eqnarray}\label{psi}
\psi(a,b)=\left\{\begin{array}{ccc}a&if&a\ge 1\\1&if&1-b\le a\textless 1\\a+b&if&a\textless 1-b\end{array}\right..\\\nn
\end{eqnarray}
For DWD loss, we have
\begin{eqnarray}\nn
\psi(a,b)=\left\{\begin{array}{ccc}a+b&if&a\le 1/2-b\\\tilde{u}&if&a\textgreater 1/2-b\end{array}\right.,\\\nn
\end{eqnarray}
where $\tilde{u}$ is the solution of the cubic equation $4u^3-4au^2-b=0$. For other loss functions, we have to rely on  certain numeric algorithms. Particularly for logistic loss, we can easily implement Newton-Raphson algorithm because the loss function has closed form second order derivatives.

Define functions $\phi_1(\cdot,\cdot,\cdot)$, $\phi_2(\cdot,\cdot,\cdot)$, and $\phi_3(\cdot,\cdot,\cdot)$ on $\mR_{\textgreater 0}\times\mR_{\textgreater 0}\times\mR_{\textgreater 0}$ as 
\begin{eqnarray}\nn
\phi_1(c_1,c_2,q)&=&E\left\{[\psi(c_1YZ_1+c_2YZ_2,q)-c_1YZ_1-c_2YZ_2]YZ_1 \right\},\\\nn
\phi_2(c_1,c_2,q)&=&E\left\{[\psi(c_1YZ_1+c_2YZ_2,q)-c_1YZ_1-c_2YZ_2]YZ_2 \right\},\\\nn
\phi_3(c_1,c_2,q)&=&E\left\{[\psi(c_1YZ_1+c_2YZ_2,q)-c_1YZ_1-c_2YZ_2]^2 \right\},
\end{eqnarray}
where  
\begin{eqnarray}\nn
Z_2\perp(Y,Z_1),~Z_1\sim N(0,1),~Z_2\sim N(0,1),\\\nn
P(Y=+1|Z_1)=g(\rho Z_1),~P(Y=-1|Z_1)=1-g(\rho Z_1).
\end{eqnarray}
We further define the asymptotic generalization error ${\cal E}^\star$ by
\begin{eqnarray}\label{precision}
{\cal E}^\star(\mu,\alpha,\tau)=P\left(\frac{R^\star}{\sqrt{q_0^\star-R^{\star 2}}}YZ\le 0\right),
\end{eqnarray}
where probability is over $Z$, $Y$ with $Z\sim N(0,1)$ and $P(Y=+1|Z)=g(\rho Z)=1-P(Y=-1|Z)$ and $q_0^\star$ and $R^\star$ are the solution of the following equations:
\begin{eqnarray}\label{eq001}
\xi_0&=&\frac{\alpha}{q^2}\phi_3\left(R,\sqrt{q_0-R^2},q\right),\\\label{eq002}
\xi&=&-\frac{\alpha \phi_2\left(R,\sqrt{q_0-R^2},q\right)}{q\sqrt{q_0-R^2}},\\\label{eq003}
\hat{R}&=&\frac{\alpha}{q}\left[\phi_1\left(R,\sqrt{q_0-R^2},q\right)-\frac{R\phi_2\left(R,\sqrt{q_0-R^2},q\right)}{\sqrt{q_0-R^2}}\right],\\\label{eq004}
q_0&=&\xi_0f_2(\xi,\tau)+\hat{R}^2f_3(\xi,\tau),\\\label{eq005}
R&=&\hat{R}f_1(\xi,\tau),\\\label{eq006}
q&=&f_0(\xi,\tau),
\end{eqnarray}
where
\begin{eqnarray}\nn
f_0(\xi,\tau)=\int\frac{X}{\xi X+\tau}\mu(dX,dW),&&f_1(\xi,\tau)=\int\frac{W^2X}{\xi X+\tau}\mu(dX,dW),\\\label{cfunc}
f_2(\xi,\tau)=\int\frac{X^2}{(\xi X+\tau)^2}\mu(dX,dW),&&f_3(\xi,\tau)=\int\frac{W^2X^2}{(\xi X+\tau)^2}\mu(dX,dW).
\end{eqnarray}

Our main mathematical results are based upon the following Proposition for the asymptotic prediction error of the estimators $\hat{\btheta}$ obtained from (\ref{class}).
\begin{proposition}\label{prop1}
Consider i.i.d. data $(\by,\bX)=\{(y_i,\bx_i)\}_{i\le n}$ where $\bx_i\sim N(0,\bSigma_n)$ and $P(y_i=+1|\bx_i)=g(\bx_i^T\btheta_{\star,n})$. Under Assumptions \ref{assum1} and \ref{assum2}, in the limit of $n,p\rightarrow\infty$ with $n/p\rightarrow\alpha$ for some positive constants $\alpha$. Let ${\cal E}_n(\by,\bX)=P(y_{new}\bx_{new}^T\hat{\btheta}(\by,\bX)\le 0)$ and ${\cal E}^\star$ be determined as per definition (\ref{precision}). Then we have, almost surely
\begin{eqnarray}\nn
\lim_{n\rightarrow\infty}{\cal E}_n(\by,\bX)\rightarrow{\cal E}^\star(\mu,\alpha,\tau).
\end{eqnarray}
\end{proposition}
The proof is given in the Appendix based on the replica method developed in statistical physics. Proposition \ref{prop1} allows us to assess the performance of different classification methods and obtain the tuning parameter value of $\tau$ that yields the maximum precision for a given method.

\subsection{Phase transition}

In this section, we derive the phase transition for the non-regularized classification methods which solve the following optimization problem 
\begin{eqnarray}\label{nonreg}
\text{argmin}_{\btheta\in{\mR}^p}\left\{\sum_{i=1}^nV(y_i\bx^T_i\btheta)\right\}.
\end{eqnarray}
A special case is that if one chooses logistic loss $V(\cdot)$, this is equivalent to the maximum likelihood estimator of logistic regression. It is well-known  that the solution of (\ref{nonreg}) does not exist in all situations, even when the number of covariates $p$ is much smaller than the sample size $n$. For instance, if the $n$ data points $(\bx_i,y_i)$ are completely linear separated in the sense that  we can find a vector $\bb\in\mR^p$ with the property $y_i\bx^T_i\bb\textgreater 0$, for all i, then the solution of (\ref{nonreg}) does not exist. If the data points overlap in the sense that for every $\bb\ne 0$, there is at least one data point satisfying $y_i\bx^T_i\bb\textgreater 0$  and at least another one satisfying $y_i\bx^T_i\bb\textless 0$, the solution of (\ref{nonreg}) does exist. Therefore, the existence for  the non-regularized classification methods undergoes a phase transition. \cite{4038449} studied the phenomenon in special case where $y_i$ is independent of $\bx_i$. This result was recently generalized by \cite{candes2020} under the significantly more challenging setting in which $P(y_i=+1|\bx_i)=1/[1+\exp(-\bx_i^T\btheta_\star)]$ and $\bx_i$ is Gaussian. Here we derive a more general result. The following Corollary allows one to characterize the minimum number of training samples per dimensions that are required in order for the non-regularized classification method (\ref{nonreg}) to have solution. 
\begin{corollary}\label{coro1}
Define $\alpha_{min}(\rho)$ as
\begin{eqnarray}\nn
1/\alpha_{min}(\rho)&=&\min_{c\in\mR}E\left\{\left(cYZ_1+Z_2\right)_+^2\right\},
\end{eqnarray}
where  $x_+=\max(x,0)$ and 
\begin{eqnarray}\nn
Z_2\perp(Y,Z_1),~Z_1\sim N(0,1),~Z_2\sim N(0,1),\\\nn
P(Y=+1|Z_1)=g(\rho Z_1),~P(Y=-1|Z_1)=1-g(\rho Z_1).
\end{eqnarray}
In the setting from Section \ref{method}, if the sample size is larger enough such that $\alpha\textgreater\alpha_{min}$, then the solution of equation (\ref{nonreg}) asymptotically exists with probability one. Conversely, if $\alpha\textless\alpha_{min}$, then the solution does not exist with probability one. 
\end{corollary}

Corollary \ref{coro1} is a generalization of the result of \cite{candes2020}, which concerns the phase transition for the existence of the maximum likelihood estimate in high-dimensional logistic regression, i.e. $g(x)$ is a logistic function. 

Note that our result is equivalent to establishing the the maximum number of training samples per dimensions below which the hard-margin SVM can have solution as shown in \cite{montanari2019generalization}. The reason is that the hard-margin SVM can only be used if the two classes in the training data are linearly separable with a positive margin. If this was not the case, the optimization problem of the hard-margin SVM would be unfeasible. Such a situation is likely to occur as a larger number of training data is used. 

For comparison, now we generalize the phase transition result for data drawn from a Gaussian mixture distribution studied in \cite{9022461}. Lets specify the joint probability distribution $P(\bx,y)$ in that scenario. Conditional on $y=\pm 1$, $\bx$ follows multivariate Gaussian distributions $P(\bx|y=\pm 1)$ with mean $\pm\bmu$ and covariance matrices $\bSigma$. Here $\bmu\in \mR^p$ and $\bSigma$ denotes the $p\times p$ positive definite matrices. From this model, we obtain the conditional distribution of $y$ given $\bx$ as
\begin{eqnarray}\nn
P(y=+1|\bx)&=&\frac{\exp\{-(\bx-\bmu)^T\bSigma^{-1}(\bx-\bmu)/2\}}{\exp\{-(\bx-\bmu)^T\bSigma^{-1}(\bx-\bmu)/2\}+\exp\{-(\bx+\bmu)^T\bSigma^{-1}(\bx+\bmu)/2\}}\\\nn
&=&\frac{1}{1+\exp(-2\bmu^T\bSigma^{-1}\bx)},
\end{eqnarray}
which is equivalent to the logistic distribution with coefficient $\btheta_\star=2\bSigma^{-1}\bmu$. The following proposition characterize the phase transition of this model in terms of 
the overall magnitude of the regression coefficient defined as $\rho^2=\btheta_\star^T\bSigma\btheta_\star=4\bmu^T\bSigma^{-1}\bmu$.
\begin{proposition}\label{prop2}
Define $\alpha_{min}(\rho)$ as the solution of
\begin{eqnarray}\nn
1&=&\alpha\int_{-\infty}^{z_c}(z_c-x)^2Dz+\left\{\alpha\rho\int_{-\infty}^{z_c}(z_c-z)Dz\right\}^2,
\end{eqnarray}
where $\Phi(z_c)=1/\alpha$ and $Dz=\frac{1}{\sqrt{2\pi}}\exp(-z^2/2)dz$. In the above Gaussian mixture setting, if the sample size per dimensions is larger enough such that $\alpha\textgreater\alpha_{min}$, then the solution of equation (\ref{nonreg}) asymptotically exists with probability one. Conversely, if $\alpha\textless\alpha_{min}$, then the solution does not exist with probability one. 
\end{proposition}
Note that Proposition \ref{prop2} generalizes the result of \cite{9022461} for hard margin SVM which can be considered as a special case here if one chooses $\bSigma=\bI_p$, where $\bI_p$ is $p$-dimensional identity matrix.

\subsection{Special examples}

In this section we illustrate our main results presented in Section \ref{main}  by considering a few special cases, namely special sequences of the true parameter vector $\btheta_{\star,n}$, and covariance matrix $\bSigma_n$. 

\subsubsection{Spiked population model}

We begin by considering data sets generated from the spiked covariance models which are particularly suitable for analyzing high dimensional statistical inference problems. Because for high dimensional data, typically only few components are scientifically important. The remaining structures can be considered as i.i.d. background noise. Therefore, we use a low-rank signal plus noise structure model \citep{ma2013,liu:sigclust}, and assume that each observation vector $\bx$ can be viewed as an independent sample from the generative models
\begin{eqnarray}\label{factor}
\bx=\sum_{k=1}^K\sqrt{\lambda_k}\bv_kz_k+\bepsilon,
\end{eqnarray} 
where $\lambda_k\textgreater 0$, $\bv_k\in \mR^p$ are orthonormal vectors, i.e. $\bv_k^T\bv_k=1$ and $\bv_k^T\bv_{k^\prime}=0$ for $k\ne k^\prime$. The random variables $z_1,\cdots,z_K$ are i.i.d N(0,1). The elements of the p-vector $\bepsilon=\{\epsilon_1,\cdots,\epsilon_p\}$ are i.i.d $N(0,1)$ which are independent of $z_k$. In model (\ref{factor}), $\lambda_k$ represents the strength of the $k$-th signal component. The real signal is typically low-dimensional, i.e. $K\ll p$. Note that the eigenvalue $\lambda_k$ is not necessarily decreasing in $k$ and $\lambda_1$ is not necessarily the largest eigenvalue. From (\ref{factor}), the covariance matrix becomes
\begin{eqnarray}\label{covariance}
\bSigma=\bI_p+\sum_{k=1}^K\lambda_k\bv_k\bv_k^T.
\end{eqnarray} 
The $k$-th eigenvalue of $\bSigma$ is $1+\lambda_k$ for $k=1,\cdots,K$ and $1$ for $k=K+1,\cdots,p$.  

Denote the projections of $\btheta_\star$ on eigenvectors as $R_k=\bv_k^T\btheta_\star$ for $k=1,\cdots,K$; $R_{K+1}=\sqrt{1-\sum_{k=1}^KR_k^2}$; and $R_k=0$ for $k=K+2,\cdots,p$. Substituting into (\ref{cfunc}), we have
\begin{eqnarray}\nn
f_0(\xi,\tau)=\frac{1}{\xi+\tau},&&f_1(\xi,\tau)=\frac{1}{\sum_{k=1}^{K+1}(1+\lambda_k)R_k^2}\sum_{k=1}^{K+1}\frac{(1+\lambda_k)^2R_k^2}{(1+\lambda_k)\xi+\tau},\\\nn
f_2(\xi,\tau)=\frac{1}{(\xi+\tau)^2},&&f_3(\xi,\tau)=\frac{1}{\sum_{k=1}^{K+1}(1+\lambda_k)R_k^2}\sum_{k=1}^{K+1}\frac{(1+\lambda_k)^3R_k^2}{[(1+\lambda_k)\xi+\tau]^2}.
\end{eqnarray} 

\subsubsection{A random features model}

We next consider a special structure of $(\bSigma,\btheta_\star)$ that captures the behavior of nonlinear random feature models, i.e. two-layers neural networks with random first layer weights. Random features methods were originally studied by \cite{neal}, \cite{DBLP:journals/ml/BalcanBV06}, and \cite{NIPS2007_3182}. It was suggested in \cite{goldt2019modelling,Aubin_2019,mei2019generalization,gerace2020generalisation} that the behavior of multilayer networks can be well approximated by certain random features model. \cite{goldt2020gaussian} proved that asymptotic behavior of the random feature models is the same as an appropriately chosen Gaussian feature model. Therefore, the two-layer neural network model can be fit within our general setting.

Assume that we perform classification on a training dataset consisting of $n$ observations $\{(\bx_i,y_i); i=1,\cdots,n\}$ generated by the latent variable $\bz_i\in N(0,\bI_d)$ through the following mechanism. The features $\bx_i$ are generated according to $x_{ij}=\sigma(\bw_j^T\bz_i)$ where $\sigma:\mR\rightarrow\mR$ is a non-linear function and $\bw_j$ are $d$-dimensional vectors drawn from $N(0,\bI_d/\sqrt{d})$. The labels
$y_i\in\{+1,-1\}$ are generated according to $P(y_i=+1|\bz_i)=f_+(\bz_i^T\bbeta_\star)$, where $\bbeta_\star\sim N(0,\bI_d/\sqrt{d})$. Denote $\bW\in\mR^{p\times d}$ the matrix with row $\bw_j$, $1\le j\le p$, we have $\bx_i=\sigma(\bW\bz_i)$ which can be described as a two layers neural network with random first-layer weights $\bW$. 

Without loss of generality, we assume $E\{\sigma(Z)\}=0$ with $Z\sim N(0,1)$. According to \cite{montanari2019generalization}, the activation function can be decomposed as 
\begin{eqnarray}\nn
\sigma(u)=\gamma_1u+\gamma_\star\sigma_\perp(u),
\end{eqnarray}
where $\gamma_1=E\{Z\sigma(Z)\}$ and $\gamma_\star^2=E\{\sigma(Z)^2\}-E\{Z\sigma(Z)\}^2-E\{\sigma(Z)\}^2$. Then the above random feature model can be described as
\begin{eqnarray}\nn
x_{ij}=\gamma_1\bw_j^T\bz_i+\gamma_\star\xi_{ij},~\xi_{ij}\perp\bz_i,~\xi_{ij}\sim N(0,1),\\\nn
g_i=\bz_i^T\bbeta_\star,~P(y_i=+1|g_i)=f_+(g_i).
\end{eqnarray}
Note that under this model $\bx_i$ and $g_i$ are jointly Gaussian with $\bx_i\sim N(0,\bSigma)$, and conditional on $\bx_i$, $g_i$ is normal with mean $\gamma_1\bbeta_\star^T\bW^T\bSigma^{-1}\bx_i$ and variance $\bbeta_\star^T\bbeta_\star-\gamma_1^2\bbeta_\star^T\bW^T\bSigma^{-1}\bW\bbeta_\star$, where $\bSigma=\gamma_1^2\bW\bW^T+\gamma^2_\star\bI_p$. For sign activation function $y_i=\text{sign}(g_i)$, $\gamma_1=\sqrt{2/\pi}$ and $\gamma_\star=\sqrt{1-2/\pi}$, we have 
\begin{eqnarray}\label{probmodel}
f_+(g_i)=P(\text{sign}(g_i)=+1)=E(g_i\ge 0)=\Phi(\bx_i^T\btheta_\star/\tilde{\tau}),
\end{eqnarray}
where $\btheta_\star=\gamma_1\bSigma^{-1}\bW\bbeta_\star$, $\tilde{\tau}^2=\bbeta_\star^T\bbeta_\star-\gamma_1^2\bbeta_\star^T\bW^T\bSigma^{-1}\bW\bbeta_\star$, and $\Phi(\cdot)$ denotes the standard Gaussian distribution function. By Marchenko-Pastur's law, the empirical spectral distribution of $\bW\bW^T$ converges to $\mu_s$ almost surely as $p,d\rightarrow\infty$ with $p/d\rightarrow\psi_1$, where
\begin{eqnarray}\nn
\mu_s(dx)&=&\left\{\begin{array}{ccc}(\psi_1-1)\delta_0+\nu_{1/\psi_1}(x)dx & if &\psi_1\ge 1\\
\nu_{\psi_1}(x)dx &if&\psi_1\in(0,1],\end{array}\right.\\\nn
\nu_\lambda&=&\frac{\sqrt{(\lambda_+-x)(x-\lambda_-)}}{2\pi\lambda x},\\\nn
\lambda_\pm&=&(1\pm\sqrt{\lambda})^2.
\end{eqnarray}
Denote the decomposition of $\bW$ as $\bW=\sum_{i=1}^p\sqrt{s_i}\bv_i\bu_i^T$, where the orthonormal vectors $\bv\in\mR^p$ and $\bu\in\mR^d$. Then we have $\bSigma=\sum_{i=1}^p\lambda_i\bv_i\bv_i^T$ with $\lambda_i=\gamma_1^2s_i+\gamma_\star^2$. According to the definition of $\rho^2=\btheta_\star^T\bSigma\btheta_\star$ and $w_i=\sqrt{p\lambda_i}\bv_i^T\btheta_\star/\rho$, we can derive
\begin{eqnarray}\nn
\rho^2=\gamma_1^2\bbeta_\star^T\bW^T\bSigma^{-1}\bW\bbeta_\star=\sum_{i=1}^p\frac{\gamma_1^2s_i(\bu_i^T\bbeta_\star)^2}{\gamma_1^2s_i+\gamma_\star^2}\rightarrow\psi_1E\frac{\gamma_1^2\tilde{X}}{\gamma_1^2\tilde{X}+\gamma_\star^2},\\\nn
w_i=\sqrt{p\lambda_i}\gamma_1\frac{\sqrt{s_i}(\bu_i^T\bbeta_\star)}{\rho\lambda_i}\rightarrow\frac{\gamma_1\sqrt{\psi_1\tilde{X}}Z}{\rho(\gamma_1^2\tilde{X}+\gamma_\star^2)^{1/2}},\\\nn
\tilde{\tau}^2\rightarrow 1-\psi_1E\frac{\gamma_1^2\tilde{X}}{\gamma_1^2\tilde{X}+\gamma_\star^2}=1-\rho^2,
\end{eqnarray}
where $\tilde{X}\sim\mu_s$ independent of $Z\sim N(0,1)$. Then the joint distribution of $\lambda,w$ converges to Law$(X,W)$, where 
\begin{eqnarray}\nn
X=\gamma_1^2\tilde{X}+\gamma_\star^2,&&W=\frac{\gamma_1\sqrt{\psi_1\tilde{X}}Z}{\rho(\gamma_1^2\tilde{X}+\gamma_\star^2)^{1/2}}.
\end{eqnarray}

\section{Numerical analysis}\label{numeric} 

In this section, we apply the general theoretical results derived in Section \ref{main} to three specific classification methods PLR, SVM, and DWD by numerically solving the nonlinear equations (\ref{eq001})-(\ref{eq006}) using the corresponding loss functions. The performance of a classification method is measured in terms of test error where the probability is over a fresh data point. Our theoretical results are verified using numerical simulations under finite size system. We aim to exploring and comparing different types of classifiers under various settings.  One main goal is to provide some guidelines on how to optimally choose classifiers and tuning parameters for a given dataset in practice. In Section \ref{phase}, we present the phase transition boundary for the separability of two classes under several settings. Then we compare the test errors of three classification methods under spiked population model in Section \ref{spike} and two layer neural network model in Section \ref{twonn}.

\subsection{Phase transition}\label{phase}

\begin{figure}[hbtp]
    \begin{center}
    \epsfig{file=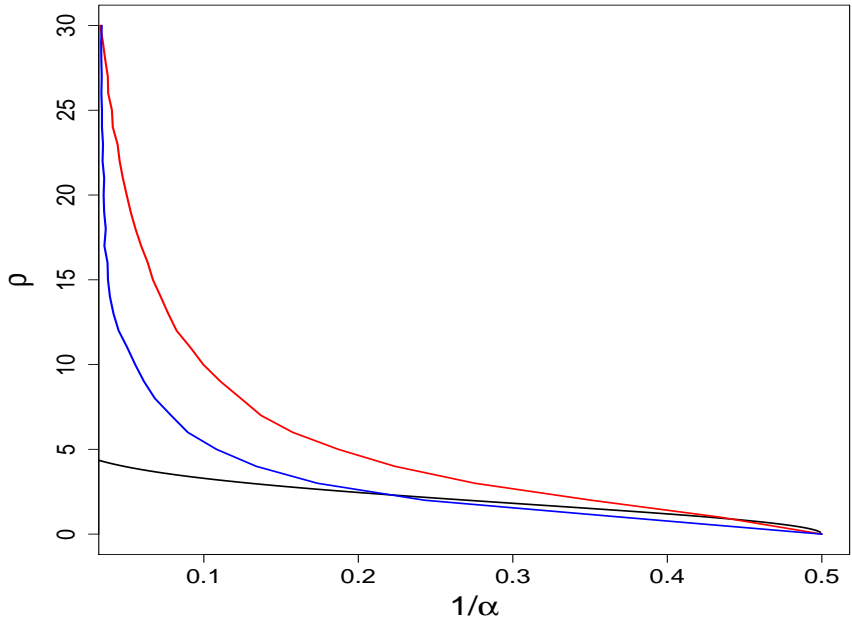,width=8.5cm,height=11.5cm,angle=-90}
    \end{center}
    \caption{Theoretical prediction for the phase transition curves. The black curve represents the boundary for Gaussian mixture model. The blue and red curves represent the boundaries for single Gaussian model with the distribution functions being probit and logit respectively.}
    \label{figure1}
\end{figure}

Figure \ref{figure1} displays the phase transition boundaries in the plane of $\rho$ and $1/\alpha$ for the separability of the two classes under different settings. Above the curve is the region where the probability of separating the two classes tends to one and below is the region where the probability of separating the two classes tends to zero. It can be seen that under the same $\alpha$, single Gaussian model needs larger $\rho$ value in order to be separated than the two Gaussian mixture model. This indicates that the data generated from a two Gaussian mixture model are easier to be separated than from a single Gaussian model. For the single Gaussian model, the data generated based on a probit distribution is easier to be separated than the data generated based on a logit distribution.

\subsection{Spiked population model}\label{spike}

\begin{figure}[hbtp]
    \begin{center}
    \epsfig{file=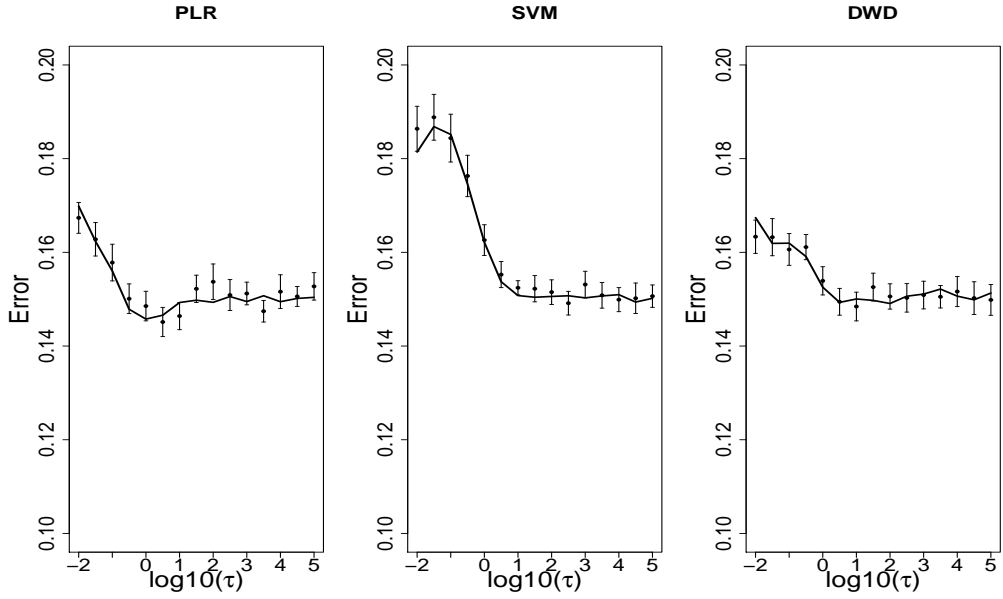,width=8.5cm,height=13.5cm,angle=-90}
    \end{center}
    \caption{Dependence of generalization error on the tuning parameter $\tau$ for different methods under spiked population model. Here $\alpha=2$ and the number of spikes $K=2$. The two spiked egenvalues $\lambda_1=\lambda_2=4$. The two projections $R_1=1/\sqrt{2}$ and $R_2=0$. The simulations are based on 20 samples with dimension $p=300$. }
    \label{figure2.1}
\end{figure}

To examine the validity of our analysis and to determine the finite-size effect, we first present some Monte Carlo simulations to confirm that our theoretical estimation derived in Section \ref{perf} is reliable. Figures \ref{figure2.1} plots the test error as a function of tuning parameter $\tau$. The comparison between our asymptotic estimations and simulations on finite dimensional datasets are also provided. We use the R packages $kernlab$, $glmnet$, and $DWDLargeR$ for solving SVM, PLR, and DWD classification problem respectively. Here the dimension of the simulated data $p=300$ and the data are generated according to (\ref{factor}) for spiked population model with i.i.d standard normal noise. We repeat simulation 20 times for each parameter setting. The mean and standard errors over 20 replications are presented. From Figures \ref{figure2.1}, we can see that our analytical curves show fairly good agreement with the simulation experiment. Thus our analytical formula (\ref{precision}) provides reliable estimates for average precision even under moderate system sizes. 

\begin{figure}[hbtp]
    \begin{center}
    \epsfig{file=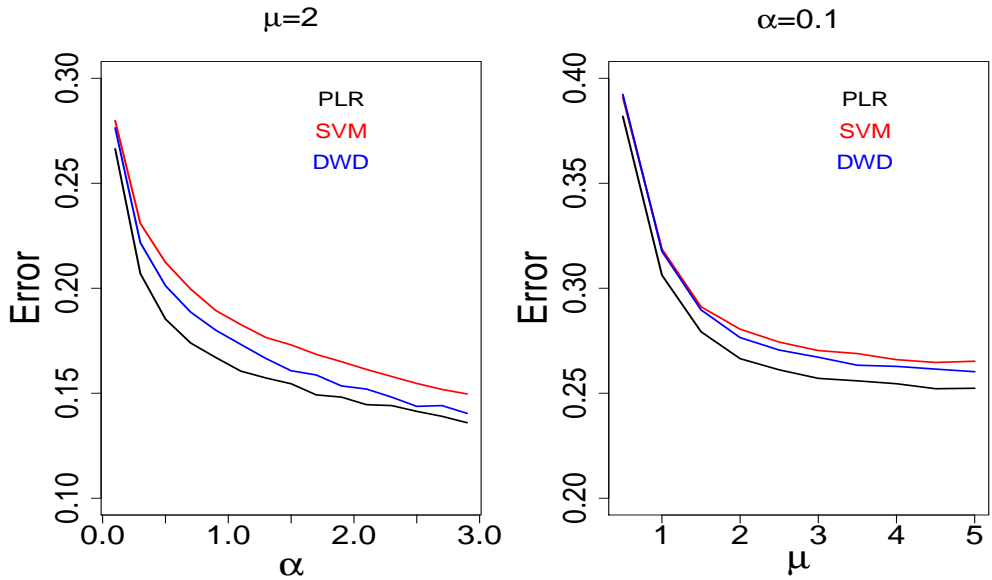,width=8.5cm,height=13.5cm,angle=-90}
    \end{center}
    \caption{Performance comparison of three classifiers at optimal tuning $\tau$ under spiked population model. Here the number of spikes $K=2$. The two spiked egenvalues $\lambda_1=\lambda_2=4$. The two projections $R_1=1/\sqrt{2}$ and $R_2=0$.}
    \label{figure2.2}
\end{figure}

Figure \ref{figure2.2} compares the performance of three classification methods after optimally tuning the parameter $\tau$. The left panel represents the dependence on $\alpha$ with $\mu$ fixed while the right panel represents the dependence on $\mu$ with $\alpha$ fixed. In both cases, PLR performs the best and SVM performs the worst while DWD is in between.

The settings of Figure \ref{figure2.1} and Figure \ref{figure2.2} are quite general in such that the spike vectors $\bv_k$ $(k=1,\cdots,K)$ are neither aligned with nor orthogonal to the signal vector $\btheta_\star$. 

\subsection{Two layer neural network model}\label{twonn}

\begin{figure}[hbtp]
    \begin{center}
    \epsfig{file=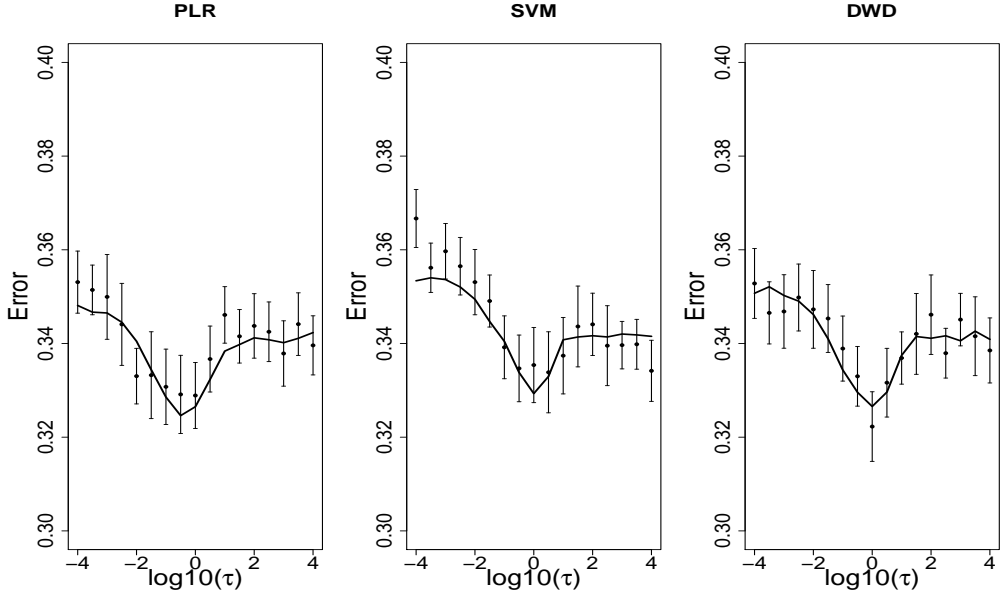,width=8.5cm,height=13.5cm,angle=-90}
    \end{center}
    \caption{Dependence of generalization error on the tuning parameter $\tau$ for different methods under the two layer neural network  model. Here $\psi_1=p/d=1$, $\psi_2=n/d=3$. The simulations are based on 20 samples with $d=200$. Sign activation function is used thus $\gamma_1=\sqrt{2/\pi}$ and $\gamma_\star=\sqrt{1-2/\pi}$.}
    \label{figure2}
\end{figure}

Figure \ref{figure2} shows the dependence of generalization error on the tuning parameter $\tau$ for two layer neural network model. The comparisons with numerical simulations are also included. The results show a fairly good agreement between theoretical prediction and Monte Carlo simulations which indicates the correctness of our analytical derivation. 

\begin{figure}[hbtp]
    \begin{center}
    \epsfig{file=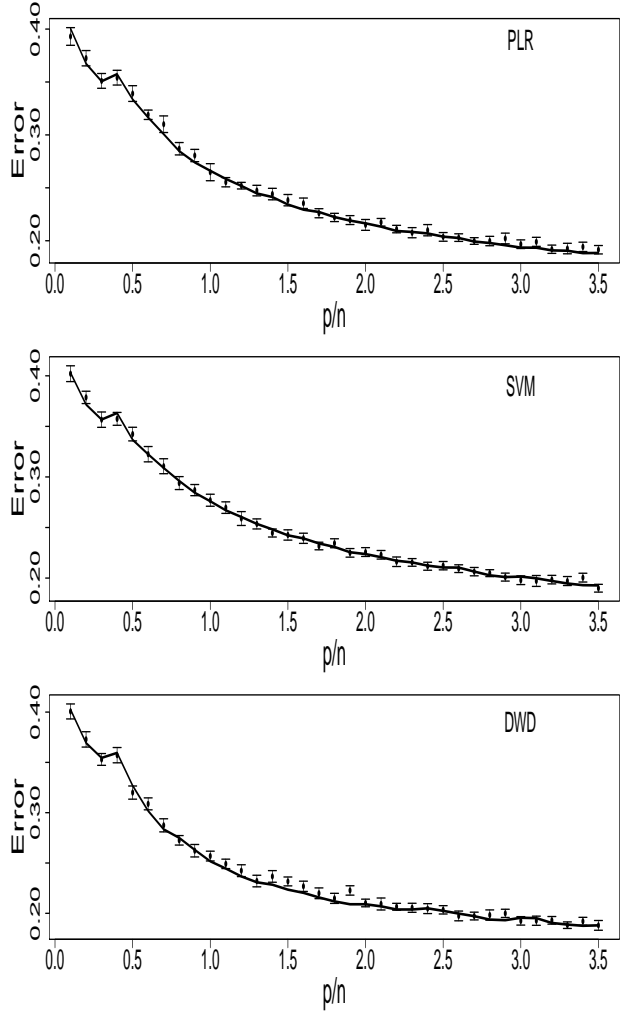,width=13.5cm,height=8.5cm,angle=-90}
    \end{center}
    \caption{Generalization error plotted against the number of features per sample at small tuning parameter $\tau=10^{-4}$. Here $\psi_2=n/d=3$. The simulations are based on 20 samples with $d=200$. Sign activation function is used thus $\gamma_1=\sqrt{2/\pi}$ and $\gamma_\star=\sqrt{1-2/\pi}$.}
    \label{figure3}
\end{figure}

In Figure \ref{figure3}, we plot the value of the generalization error as a function of $p/n$  with fixed $\psi_2=n/d$ at small values of the regularization parameter $\tau=10^{-4}$. We show the so-called double descent behavior for all three classification methods with a peak at the threshold value where the data become linearly separable. This finding agrees with the recently developed `double descent' phenomenology for hard margin SVM in \cite{montanari2019generalization} and logistic regression in \cite{goldt2019modelling}. 

\begin{figure}[hbtp]
    \begin{center}
    \epsfig{file=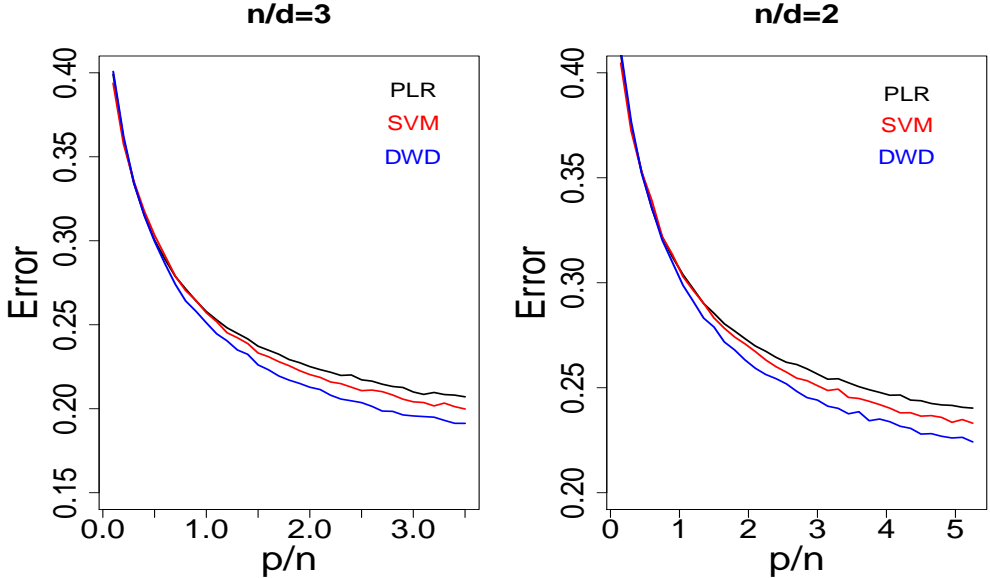,width=8.5cm,height=13.5cm,angle=-90}
    \end{center}
    \caption{Performance comparison of three classifiers at optimal tuning $\tau$ under the two layer neural network model. Sign activation function is used thus $\gamma_1=\sqrt{2/\pi}$, $\gamma_\star=\sqrt{1-2/\pi}$.}
    \label{figure4}
\end{figure}

Figure \ref{figure4} compares the performance of three classification methods after optimally tuning the parameter $\tau$ for two layer neural network model. For two fixed ratios between the number of samples and dimension $d$, the generalization errors of three methods are very close at small value of overparametrization ratio $p/n$.  For large $p/n$, DWD performs the best and PLR performs the worst while SVM is in between. This is different from the performance under the spiked population model as shown in Figure \ref{figure2.2}.

\section{Conclusion}\label{discussion}

Large margin classifiers are commonly used in practice. In this paper, we examine the limiting behavior of a general family of large-margin classifiers as $p,n\rightarrow\infty$ with fixed $\alpha=n/p$. This family is very general and it includes many popular classification methods as special cases. We illustrate our main results by considering two special covariance structures: spiked population model and two layer neural network model with random first layer weights. We explore the phase transition behavior for the separability of  the two classes and our general conclusion covers several existing results as special cases. Our results can provide some practical guidelines for selecting the best model as well as the optimal tuning parameter for a given classification problem.  Although our theoretical results are asymptotic in the problem dimensions, numerical simulations have shown that they are accurate already on problems with a few hundreds of variables. Our main observations from the derived analytic formulas are
\begin{itemize}
\item Under the same condition, data generated from Gaussian mixture distribution are easier to be separated than from single Gaussian distribution. 
\item For spiked population covariance structure, after optimally tuning the regularization parameter, PLR yields the best classification performance, followed by DWD and SVM. 
\item For two layer neural network covariance structure, after optimally tuning the regularization parameter, the three methods almost yields the same classification performance when $p/n$ is small. However, at large value of $p/n$, DWD yields the best classification performance, followed by PLR and SVM.
\item For two layer neural network covariance structure, we reproduce the double descent phenomenon for all three methods. We show that the test error peaks at a critical value of $\psi_1$ when the two classes become separable. 
\end{itemize}
It is interesting to note that our findings provide theoretical confirmations to the empirical results observed in \cite{Marron2007} that DWD yields superior performance to SVM in HDLSS situations. This statement has been confirmed in \cite{huang2019large} for the Gaussian mixture model. Here it is also confirmed to be true for the single Gaussian model. Although our observations may not hold for all covariance structure, it can help us to understand the classification behaviors of different methods better.

%
%
%


\section*{Appendix}
\label{sec1}
\setcounter{equation}{0}
\def\theequation{A\arabic{equation}}

\section*{Proof of Proposition \ref{prop1}}
This appendix outlines the replica calculation leading to Propositions 1. We limit ourselves to the main steps. For a general introduction to the method and its motivation, we refer to \cite{mezard1987spin,mezard2009information,PhysRevX.2.021005}. 

Denote $\bX=[\bx_1,\cdots,\bx_n]^T$, $\by=(y_1,\cdots,y_n)^T$. We consider regularized classification of the form
\begin{eqnarray}\label{classreg}
\hat{\btheta}&=&\text{argmin}_{\btheta}\left\{\sum_{i=1}^nV\left(\frac{y_i\bx_i^T\btheta}{\sqrt{p}}\right)+\sum_{j=1}^pJ_\tau(\theta_j)\right\}.
\end{eqnarray}
After suitable scaling, the terms inside the bracket $\{\cdot\}$ are exactly equal to the objective function of model (\ref{class}) in the main text. 

The replica calculation aims at estimating the following moment generating function (partition function)
\begin{eqnarray}\nn
&&Z_{\beta}(\bX,\by)\\\label{zbeta}
&=&\int\exp\left\{-\beta\left[\sum_{i=1}^nV\left(\frac{y_i\bx_i^T\btheta}{\sqrt{p}}\right)+\sum_{j=1}^pJ_\tau(\theta_j)\right]\right\}d\btheta
\end{eqnarray}
where $\beta\textgreater 0$ is a `temperature' parameter. In the zero temperature limit, i.e. $\beta\rightarrow\infty$, $Z_{\beta}(\bX,\by)$ is dominated by the values of $\btheta$ which are the solution of (\ref{classreg}). 

Within the replica method, it is assumed that the limits $p\rightarrow\infty$, $\beta\rightarrow\infty$ exist almost surely for the quantity $(p\beta)^{-1}\log Z_{\beta}(\bX,\by)$, and that the order of the limits can be exchanged. We therefore define the free energy 
\begin{eqnarray}\nn
{\cal F}&=&-\lim_{\beta\rightarrow\infty}\lim_{p\rightarrow\infty}\frac{1}{p\beta}\log Z_{\beta}(\bX,\by)=-\lim_{p\rightarrow\infty}\lim_{\beta\rightarrow\infty}\frac{1}{p\beta}\log Z_{\beta}(\bX,\by).
\end{eqnarray}
It is also assumed that $p^{-1}\log Z_{\beta}(\bX,\by)$ concentrates tightly around its expectation so that the free energy can in fact be evaluated by computing
\begin{eqnarray}\label{calf}
{\cal F}&=&-\lim_{\beta\rightarrow\infty}\lim_{p\rightarrow\infty}\frac{1}{p\beta}\left\langle\log Z_{\beta}(\bX,\by)\right\rangle_{\bX,\by},
\end{eqnarray}
where the angle bracket stands for the expectation with respect to the distribution of training data $\bX$ and $\by$. Notice that, by (\ref{calf}) and using Laplace method in the integral (\ref{zbeta}), we have
\begin{eqnarray}\nn
{\cal F}&=&\lim_{p\rightarrow\infty}\frac{1}{p}\min_{\btheta}\left\{\sum_{i=1}^nV\left(\frac{y_i\bx_i^T\btheta}{\sqrt{p}}\right)+\sum_{j=1}^pJ_\tau(\theta_j)\right\}.
\end{eqnarray}
  
In order to evaluate the integration of a log function, we make use of the replica method based on the identity 
\begin{eqnarray}
\log Z=\lim_{k\rightarrow 0}\frac{\partial Z^k}{\partial k}=\lim_{k\rightarrow 0}\frac{\partial}{\partial k}\log Z^k,
\end{eqnarray}
and rewrite (\ref{calf}) as
\begin{eqnarray}\label{replica}
{\cal F}=-\lim_{\beta\rightarrow\infty}\lim_{p\rightarrow\infty}\frac{1}{p\beta}\lim_{k\rightarrow 0}\frac{\partial}{\partial k}\log\Xi_k(\beta),
\end{eqnarray}
where 
\begin{eqnarray}\label{xi0}
\Xi_k(\beta)=\langle\{Z_\beta(\bX,\by)\}^k\rangle_{\bX,\by}=\int\{Z_\beta(\bX,\by)\}^k\prod_{i=1}^nP(\bx_i,y_i)d\bx_i dy_i.
\end{eqnarray}
Equation (\ref{replica}) can be derived by using the fact that $\lim_{k\rightarrow 0}\Xi_k(\beta)=1$ and exchanging the order of 
the averaging and the differentiation with respect to $k$. In the replica method, we will first evaluate $\Xi_k(\beta)$ for integer $k$ and then 
apply to real $k$ and take the limit of $k\rightarrow 0$. 

For integer $k$, in order to represent $\{Z_\beta(\bX,\by)\}^k$ in the integrand of (\ref{xi0}), we use the identity
\begin{eqnarray}\nn
\left(\int f(x)\mu(dx)\right)^k=\int f(x_1)\cdots f(x_k)\mu(dx_1)\cdots\mu(dx_k),
\end{eqnarray}
and obtain
\begin{eqnarray}\label{zd}
\{Z_\beta(\bX,\by)\}^k&=&\prod_{a=1}^k\left[\int \exp\left\{-\beta\left[\sum_{i=1}^nV\left(\frac{y_i\bx_i^T\btheta^a}{\sqrt{p}}\right)+\sum_{j=1}^pJ_\tau(\theta_j^a)\right]\right\}d\btheta^a\right]
\end{eqnarray}
where we have introduced replicated parameters
\begin{eqnarray}\nn
\btheta^a\equiv[\theta^a_1,\cdots,\theta^a_p]^T, \text{ for }a=1,\cdots,k.
\end{eqnarray}
Exchanging the order of the two limits $p\rightarrow\infty$ and $k\rightarrow 0$ in (\ref{replica}), we have 
\begin{eqnarray}\label{freef}
{\cal F}=-\lim_{\beta\rightarrow\infty}\frac{1}{\beta}\lim_{k\rightarrow 0}\frac{\partial}{\partial k}\left(\lim_{p\rightarrow\infty}\frac{1}{p}\log\Xi_k(\beta)\right).
\end{eqnarray}

Define the measure $\nu(d\btheta)$ over $\btheta\in\mathbb{R}^p$ as follows
\begin{eqnarray}\nn
\nu(d\btheta)&=&\exp\left\{-\beta\sum_{j=1}^pJ_\tau(\theta_j)\right\}d\btheta.
\end{eqnarray}
Similarly, define the measure $\nu(d\bx)$ as $\nu(d\bx)=P(\bx)d\bx$. In order to carry out the calculation of $\Xi_k(\beta)$, we let $\nu^k(d\btheta)\equiv\nu(d\btheta^1)\times\cdots\times\nu(d\btheta^k)$ be a measure over $(\mathbb{R}^p)^k$, with $\btheta^1,\cdots,\btheta^k\in\mathbb{R}^p$. Analogously $\nu^n(d\bx)\equiv\nu(d\bx_1)\times\cdots\times\nu(d\bx_n)$ with $\bx_1,\cdots,\bx_n\in\mathbb{R}^p$ and $\nu^n(dy)=\nu(dy_1)\cdots\nu(dy_n)$. With these notations, we have
\begin{eqnarray}\nn
\Xi_k(\beta)&=&\int\exp\left\{-\beta\sum_{i=1}^n\sum_{a=1}^kV\left(\frac{y_i\bx_i^T\btheta^a}{\sqrt{p}}\right)\right\}\nu^k(d\btheta)\nu^n(d\bx)\nu^n(dy)\\\label{xi}
&=&\int\{ I(\btheta)\}^{n}\nu^k(d\btheta),\\\nn
\end{eqnarray}
where 
\begin{eqnarray}\nn
&&I(\btheta)\\\label{iterm}
&=&\int\int\exp\left\{-\beta\sum_{a=1}^kV\left(\frac{y\bx^T\btheta^a}{\sqrt{p}}\right)\right\}\nu(d\bx)\nu(dy)\\\nn
&=&\int\left[\exp\left\{-\beta\sum_{a=1}^kV\left(\frac{\bx^T\btheta^a}{\sqrt{p}}\right)\right\}f_+(\frac{\bx^T\btheta_\star}{\sqrt{p}})+\exp\left\{-\beta\sum_{a=1}^kV\left(\frac{-\bx^T\btheta^a}{\sqrt{p}}\right)\right\}f_-(\frac{\bx^T\btheta_\star}{\sqrt{p}})\right]\nu(d\bx),
\end{eqnarray}
where $f_+(\frac{\bx^T\btheta_\star}{\sqrt{p}})=\Phi(\bx_i^T\btheta_\star/\tilde{\tau})$ and $f_-(\frac{\bx^T\btheta_\star}{\sqrt{p}})=\Phi(-\bx_i^T\btheta_\star/\tilde{\tau})$ as shown in (\ref{probmodel}).
Notice that above we used the fact that the integral over $(\bx_1,\cdots,\bx_n)\in(\mathbb{R}^p)^n$ factors into $n$ integrals over $(\mathbb{R})^p$ with measure $\nu(d\bx)$. We next use the identity 
\begin{eqnarray}\label{identity}
f(x)&=&\frac{1}{2\pi}\int^{\infty}_{-\infty}\int^{\infty}_{-\infty}f(q)e^{i\left(q-x\right)\hat{q}}dqd\hat{q}.
\end{eqnarray}
We apply this identity to (\ref{iterm}) and introduce integration variables $du^a,d\hat{u}^a$ for $1\le a\le k$. Letting $\nu^k(du)=du^1\cdots du^k$ and $\nu^k(d\hat{u})=d\hat{u}^1\cdots d\hat{u}^k$
\begin{eqnarray}\nn
I(\btheta)&=&\int\left[\exp\left\{-\beta\sum_{a=1}^kV(u^a)\right\}f_+(u^\star)+\exp\left\{-\beta\sum_{a=1}^kV(-u^a)\right\}f_-(u^\star)\right]\\\nn
&&\exp\left\{i\sqrt{p}\sum_{a=1}^k\left(u^a-\frac{\bx^T\btheta^a}{\sqrt{p}}\right)\hat{u}^a+i\sqrt{p}\left(u^\star-\frac{\bx^T\btheta_\star}{\sqrt{p}}\right)\hat{u}^\star\right\}\nu(d\bx)\nu^k(du)\nu^k(d\hat{u})\nu(du^\star)\nu(d\hat{u}^\star)\\\nn
&=&\int\left[\exp\left\{-\beta\sum_{a=1}^kV(u^a)\right\}f_+(u^\star)+\exp\left\{-\beta\sum_{a=1}^kV(-u^a)\right\}f_-(u^\star)\right]\\\nn
&&\exp\left\{i\sqrt{p}\sum_{a=1}^ku^a\hat{u}^a+i\sqrt{p}u^\star\hat{u}^\star-\frac{1}{2}\sum_{ab}(\btheta^a)^T\bSigma\btheta^b\hat{u}^a\hat{u}^b\right.\\\label{iplus}
&&\left.-\frac{1}{2}(\btheta_\star)^T\bSigma\btheta_\star\hat{u}^\star\hat{u}^\star-\sum_{a}(\btheta^a)^T\bSigma\btheta_\star\hat{u}^a\hat{u}^\star\right\}\nu^k(du)\nu^k(d\hat{u})du^\star d\hat{u}^\star.
\end{eqnarray}
 In deriving (\ref{iplus}), we have used the fact that the low-dimensional marginals of $\bx$ can be approximated by Gaussian distribution based on multivariate central limit theorem.  

Next we apply (\ref{identity}) to (\ref{xi}), and introduce integration variables $Q_{ab},\hat{Q}_{ab}$ and $R^a,\hat{R}^a$ associated with $(\btheta^a)^T\bSigma\btheta^b/p$ and $(\btheta^a)^T\bSigma\btheta_\star/p$ respectively for $1\le a,b\le k$. Denote $\bQ\equiv(Q_{ab})_{1\le a,b\le k}$, $\hat{\bQ}\equiv(\hat{Q}_{ab})_{1\le a,b\le k}$, $\bR\equiv(R^a)_{1\le a\le k}$, and $\hat{\bR}\equiv(\hat{R}^a)_{1\le a\le k}$. Note that, constant factors can be applied to the integration variables, and we choose convenient factors for later calculations.  Letting $d\bQ\equiv\prod_{a,b}dQ_{ab}$, $d\hat{\bQ}\equiv\prod_{a,b}d\hat{Q}_{ab}$, $d\bR\equiv\prod_{a}dR^{a}$, and $d\hat{\bR}\equiv\prod_{a}d\hat{R}^{a}$, we obtain
\begin{eqnarray}\nn
\Xi_k(\beta)&=&\int\{\hat{\xi}(\bQ,\bR)\}^n\exp\left\{i\sum_{ab}pQ_{ab}\hat{Q}_{ab}+i\sum_{a}pR_a\hat{R}_a-i\sum_{ab}(\btheta^a)^T\bSigma\btheta^b\hat{Q}_{ab}-i\sum_{a}(\btheta^a)^T\bSigma\btheta_\star\hat{R}_a\right\}\\\label{xik}
&&d\bQ d\hat{\bQ}d\bR d\hat{\bR}\nu^k(d\btheta),
\end{eqnarray}
where
\begin{eqnarray}\nn
\hat{\xi}(\bQ,\bR)&=&\int\left[\exp\left\{-\beta\sum_{a=1}^kV(u^a)\right\}f_+(u^\star)+\exp\left\{-\beta\sum_{a=1}^kV(-u^a)\right\}f_-(u^\star)\right]\\\nn
&&\exp\left\{i\sqrt{p}\sum_{a=1}^ku^a\hat{u}^a+i\sqrt{p}u^\star\hat{u}^\star-\frac{1}{2}\sum_{ab}pQ_{ab}\hat{u}^a\hat{u}^b-\frac{1}{2}p\rho^2\hat{u}^\star\hat{u}^\star-\sum_{a}pR^a\hat{u}^a\hat{u}^\star\right\}\\\label{ihat}
&&\nu^k(du)\nu^k(d\hat{u})du^\star d\hat{u}^\star.
\end{eqnarray}
Now we can rewrite (\ref{xik}) as
\begin{eqnarray}\label{saddle}
\Xi_k(\beta)&=&\int\exp\left\{-p{\cal S}_k(\bQ,\hat{\bQ},\bR,\hat{\bR})\right\}d\bQ d\hat{\bQ}d\bR d\hat{\bR},
\end{eqnarray}
where
\begin{eqnarray}\nn
{\cal S}_k(\bQ,\hat{\bQ},\bR,\hat{\bR})&=&-i\beta\left(\sum_{ab}Q_{ab}\hat{Q}_{ab}+\sum_aR^{a}\hat{R}^{a}\right)-\frac{1}{p}\log\xi(\hat{\bQ},\hat{\bR})-\alpha\log\hat{\xi}(\bQ,\bR),\\\label{xihat}
\xi(\hat{\bQ},\hat{\bR})&=&\int\exp\left\{-i\sum_{ab}\hat{Q}_{ab}(\btheta^a)^T\bSigma\btheta^b-i\sum_{a}(\btheta^a)^T\bSigma\btheta_\star\hat{R}_a\right\}\nu^k(d\btheta).
\end{eqnarray}
Now we apply steepest descent method to the remaining integrations. According to Varadhan's proposition \citep{Tanaka}, only the saddle points of the exponent of the integrand contribute to the integration in the limit of $p\rightarrow\infty$. We next use the saddle point method in (\ref{saddle}) to obtain
\begin{eqnarray}\nn
-\lim_{p\rightarrow\infty}\frac{1}{p}\Xi_k(\beta)&=&{\cal S}_k(\bQ^\star,\hat{\bQ}^\star,\bR^\star,\hat{\bR}^\star),
\end{eqnarray}
where $\bQ^\star,\hat{\bQ}^\star,\bR^\star,\hat{\bR}^\star$ are the saddle point location. Looking for saddle-points over all the entire space is in general difficult to perform. We  assume replica symmetry for saddle-points such that they are invariant under exchange of any two replica indices $a$ and $b$, where $a\ne b$. Under this symmetry assumption, the space is greatly reduced and the exponent of the integrand can be explicitly evaluated.  The replica symmetry is also motivated by the fact that ${\cal S}_k(\bQ^\star,\hat{\bQ}^\star,\bR^\star,\hat{\bR}^\star)$ is indeed left unchanged by such change of variables. This is equivalent to postulating that $R^a=R$, $\hat{R}^a=i\hat{R}$, 
\begin{eqnarray}\label{qsym}
(Q_{ab})^\star=\left\{\begin{array}{cc}q_1&\text{if a=b}\\q_0&\text{otherwise}\end{array}\right.,&\text{and}&(\hat{Q}_{ab})^\star=\left\{\begin{array}{cc}i\frac{\beta\xi_1}{2}&\text{if a=b}\\i\frac{\beta\xi_0}{2}&\text{otherwise}\end{array}\right.,
\end{eqnarray}
where the factor $i\beta/2$ is for future convenience. The next step consists in substituting the above expressions for $\bQ^\star,\hat{\bQ}^\star,\bR^\star,\hat{\bR}^\star$ in ${\cal S}_k(\bQ^\star,\hat{\bQ}^\star,\bR^\star,\hat{\bR}^\star)$ and then taking the limit $k\rightarrow 0$. We will consider separately each term of ${\cal S}_k(\bQ^\star,\hat{\bQ}^\star,\bR^\star,\hat{\bR}^\star)$. Let us begin with the first term
\begin{eqnarray}\label{qqhat}
-i\beta\left(\sum_{ab}Q_{ab}\hat{Q}_{ab}+\sum_aR^a\hat{R}^{a}\right)&=&\frac{k\beta^2}{2}(\xi_1q_1-\xi_0q_0)+k\beta R\hat{R}.
\end{eqnarray}
Let us consider $\log\xi(\hat{\bQ},\hat{\bR})$. For p-vectors $\bu,\bv\in \mR^p$ and $p\times p$ matrix $\bSigma$, introducing the notation $\|\bv\|_{\bSigma}^2\equiv \bv^T\bSigma\bv$ and $\langle\bu,\bv\rangle\equiv\sum_{j=1}^pu_jv_j/p$, we have
\begin{eqnarray}\nn
\xi(\hat{\bQ},\hat{\bR})&=&\int\exp\left\{\frac{\beta^2}{2}(\xi_1-\xi_0)\sum_{a=1}^k\|\btheta^a\|^2_{\bSigma}+\frac{\beta^2\xi_0}{2}\sum_{a,b=1}^k(\btheta^a)^T\bSigma\btheta^b\right.\\\nn
&&~~~~~~~~~\left.+\beta\sum_{a=1}^k\hat{R}(\btheta^a)^T\bSigma\btheta_\star\right\}\nu^k(d\btheta)\\\nn
&=&E\int\exp\left\{\frac{\beta^2}{2}(\xi_1-\xi_0)\sum_{a=1}^k\|\btheta^a\|^2_{\bSigma}+\beta\sqrt{\xi_0}\sum_{a=1}^k(\btheta^a)^T\bSigma^{1/2}\bz\right.\\\label{xiq}
&&~~~~~~~~~\left.+\beta\sum_{a=1}^k\hat{R}(\btheta^a)^T\bSigma\btheta_\star\right\}\nu^k(d\btheta),
\end{eqnarray}
where expectation is with respect to $\bz\sim N(0,\bI_{p})$. Notice that, given $\bz\in \mR^p$, the integrals over $\btheta^1,\cdots,\btheta^k$ factorize, whence
\begin{eqnarray}\nn
\xi(\hat{\bQ},\hat{\bR})&=&E\left\{\left[\int\exp\left\{\frac{\beta^2}{2}(\xi_1-\xi_0)\|\btheta\|^2_{\bSigma}+\beta\sqrt{\xi_0}\btheta^T\bSigma^{1/2}\bz\right.\right.\right.\\\nn
&&\left.\left.\left.+\beta\hat{R}(\btheta)^T\bSigma\btheta_\star\right\}\nu(d\btheta)\right]^k\right\}.
\end{eqnarray}
Finally, after integration over $\nu^k(d\hat{u})$, (\ref{ihat}) becomes
\begin{eqnarray}\nn
\hat{\xi}(\bQ,\bR)&=&\int\left[\exp\left\{-\beta\sum_{a=1}^kV(u^a)\right\}f_+(u^\star)+\exp\left\{-\beta\sum_{a=1}^kV(-u^a)\right\}f_-(u^\star)\right]\\\nn
&&\exp\left\{i\sqrt{p}u^\star\hat{u}^\star-\frac{1}{2}p\rho^2\hat{u}^\star\hat{u}^\star-\frac{1}{2}\sum_{ab}(u^a+i\sqrt{p}R^a\hat{u}^\star)(\bQ^{-1})_{ab}(u^b+i\sqrt{p}R^b\hat{u}^\star)-\frac{1}{2}\log\text{det}\bQ\right\}\\\label{ihat}
&&\nu^k(du)du^\star d\hat{u}^\star.
\end{eqnarray}

We can next take the limit $\beta\rightarrow\infty$. The analysis of the saddle point parameters $q_0,q_1,\xi_0,\xi_1$ shows that $q_0,q_1$ have the same limit with $q_1-q_0=(q/\beta)+o(\beta^{-1})$ and $\xi_0,\xi_1$ have the same limit with $\xi_1-\xi_0=(-\xi/\beta)+o(\beta^{-1})$. Substituting the above expression in (\ref{qqhat}) and (\ref{xiq}), in the limit of $k\rightarrow 0$, we then obtain
\begin{eqnarray}\label{first}
-i\beta\left(\sum_{ab}Q_{ab}\hat{Q}_{ab}+\sum_aR^a\hat{R}^{a}\right)
&=&\frac{k\beta}{2}(\xi_0q-\xi q_0)+k\beta R\hat{R},
\end{eqnarray}
and
\begin{eqnarray}\nn
\xi(\hat{\bQ},\hat{\bR})&=&E\left\{\left[\int\exp\left\{-\frac{\beta\xi}{2}\|\btheta\|^2_{\bSigma}+\beta\sqrt{\xi_0}\btheta^T\bSigma^{1/2}\bz\right.\right.\right.\\\label{second}
&&\left.\left.\left.+\beta\hat{R}(\btheta)^T\bSigma\btheta_\star\right\}\nu(d\btheta)\right]^k\right\}.
\end{eqnarray}
Similarly, using (\ref{qsym}), we obtain
\begin{eqnarray}\nn
\sum_{ab}(u^a+i\sqrt{p}R^a\hat{u}^\star)(\bQ^{-1})_{ab}(u^b+i\sqrt{p}R^b\hat{u}^\star)
&=&\frac{\beta\sum_a(u^a+i\sqrt{p}R^a\hat{u}^\star)^2}{q}-\frac{\beta^2q_0\{\sum_a(u^a+i\sqrt{p}R^a\hat{u}^\star)\}^2}{(q)^2},\\\nn
\log\text{det}\bQ&=&\log\left[(q_1-q_0)^k\left(1+\frac{kq_0}{q_1-q_0}\right)\right]=\frac{k\beta q_0}{q},
\end{eqnarray}
where we retain only the leading order terms. Therefore, (\ref{ihat}) becomes
\begin{eqnarray}\nn
\hat{\xi}(\bQ,\bR)&=&\int\left[\exp\left\{-\beta\sum_{a=1}^kV(u^a)\right\}f_+(u^\star)+\exp\left\{-\beta\sum_{a=1}^kV(-u^a)\right\}f_-(u^\star)\right]\\\nn
&&\exp\left\{i\sqrt{p}u^\star\hat{u}^\star-\frac{1}{2}p\rho^2\hat{u}^\star\hat{u}^\star-\frac{\beta\sum_{a}(u^a)^2}{2q}-\frac{i\sqrt{p}\beta\hat{u}^\star\sum_au^aR^a}{q}+\frac{\beta^2q_0(\sum_au^a)^2}{2q^2}-\frac{k\beta q_0}{2q}\right\}\\\nn
&&\nu^k(du)\\\nn
&=&E_{u^\star}\int\left[\exp\left\{-\beta\sum_{a=1}^kV(u^a)\right\}f_+(u^\star)+\exp\left\{-\beta\sum_{a=1}^kV(-u^a)\right\}f_-(u^\star)\right]\\\nn
&&\exp\left\{-\frac{\beta\sum_{a}(u^a)^2}{2q}+\frac{\beta^2(q_0-R^2/\rho^2)(\sum_au^a)^2}{2q^2}+\frac{\beta Ru^\star\sum_au^a}{q\rho^2}-\frac{k\beta q_0}{2q}\right\}\nu^k(du)\\\nn
&=&\exp\left(-\frac{k\beta q_0}{2q}\right)E_zE_{u^\star}\\\nn
&&\left[\left\{\int\exp\left\{-\beta V(u)-\frac{\beta u^2}{2q}+\frac{\beta\sqrt{q_0-R^2/\rho^2}zu}{q}+\frac{\beta Ru^\star u}{q\rho}\right\}du\right\}^kf_+(\rho u^\star)\right.\\\nn
&+&\left.\left\{\int\exp\left\{-\beta V(-u)-\frac{\beta u^2}{2q}+\frac{\beta\sqrt{q_0-R^2/\rho^2}zu}{q}+\frac{\beta Ru^\star u}{q\rho}\right\}du\right\}^kf_-(\rho u^\star)\right]\\\nn
&=&\exp\left(-\frac{k\beta q_0}{2q}\right)E_zE_{u^\star}E_{y^\star}\left(\int\exp\left\{-\beta V(u)-\frac{\beta(u-y^\star u^\star R/\rho-\sqrt{q_{0}-R^2/\rho^2}y^\star z)^2}{2q}\right.\right.\\\nn
&&~~~~~~~~~~~~~~~~~~~~~~~~~~~~~~~~~~~~~~~~\left.\left.+\frac{\beta(\sqrt{q_{0}-R^2/\rho^2}y^\star z+y^\star u^\star R/\rho)^2}{2q}\right\}du\right)^k,
\end{eqnarray}
where the expectation $z\perp u,~z\sim N(0,1),~u^\star\sim N(0,1)$, and $P(y^\star=\pm|u^\star)=f_\pm(\rho u^\star)$. Substituting this expression in (\ref{xihat}), we obtain
\begin{eqnarray}\label{third}
\log\hat{\xi}(\bQ,\bR)&=&-k\beta E\left\{\min_u\left[V(u)+\frac{(u-y^\star u^\star R/\rho-\sqrt{q_{0}-R^2/\rho^2}y^\star z)^2}{2q}\right]\right\},
\end{eqnarray}
where the expectation is with respect to $z$, $u^\star$, and $y^\star$. Putting (\ref{first}), (\ref{second}), and (\ref{third}) together into (\ref{saddle}) and then into (\ref{replica}), we obtain 
\begin{eqnarray}\nn
{\cal F}&=&\frac{1}{2}(\xi_0q-\xi q_0)+R\hat{R}\\\nn
&&+\alpha E\left\{\min_u\left[V(u)+\frac{\left(u-y^\star u^\star R/\rho-\sqrt{q_{0}-R^2/\rho^2}y^\star z\right)^2}{2q}\right]\right\}\\\label{fs}
&&+\frac{1}{p}\text{E}\min_{\btheta\in\mR^p}\left\{\frac{\xi}{2}\|\btheta\|_{\bSigma}^2
-\left\langle\sqrt{\xi_0}\bSigma^{1/2}\bz+\hat{R}\bSigma\btheta_\star,\bw\right\rangle+\sum_{j=1}^pJ_\tau(\theta_j)\right\},
\end{eqnarray}
where the expectations are with respect to $ z$, $u^\star$, and $y^\star$. Here $\xi,~\xi_0,~q,~q_0,~R,~\hat{R}$ are order parameters which can be determined from the saddle point equations of ${\cal F}$. Define functions $\phi_1$, $\phi_2$, and $\phi_3$ as 
\begin{eqnarray}\nn
\phi_1&=&E\left\{\left(\hat{u}-y^\star u^\star R/\rho-\sqrt{q_{0}-R^2/\rho^2}y^\star z\right)y^\star u^\star\right\},\\\nn
\phi_2&=&E\left\{\left(\hat{u}-y^\star u^\star R/\rho-\sqrt{q_{0}-R^2/\rho^2}y^\star z\right)y^\star z\right\},\\\nn
\phi_3&=&E\left\{\left(\hat{u}-y^\star u^\star R/\rho-\sqrt{q_{0}-R^2/\rho^2}y^\star z\right)^2\right\},
\end{eqnarray}
where  
\begin{eqnarray}\nn
\hat{u}&=&\text{argmin}_{u\in\mR}\left\{V(u)+\frac{\left(u-y^\star u^\star  R/\rho -\sqrt{q_{0}-R^2/\rho^2}y^\star z\right)^2}{2q}\right\}.
\end{eqnarray}
The result in (\ref{fs}) is for general penalty function $J_\tau(w)$. For quadratic penalty $J_\tau(w)=\tau w^2$, we get the closed form limiting distribution of $\bw$ as 
\begin{eqnarray}\label{wlimit}
\hat{\btheta}&=&(\xi\bSigma+\tau \bI_p)^{-1}\left(\sqrt{\xi_0}\bSigma^{1/2}\bz+\hat{R}\bSigma\btheta_\star\right).
\end{eqnarray}
All the order parameters can be determined by the following saddle-point equations:
\begin{eqnarray}\label{aeq001}
\xi_0&=&\frac{\alpha}{q^2}\phi_3,\\\label{aeq002}
\xi&=&-\frac{\alpha\phi_2}{q\sqrt{q_0-R^2/\rho^2}},\\\label{aeq003}
\hat{R}&=&\frac{\alpha}{q}\left(\frac{\phi_1}{\rho}-\frac{R\phi_2}{\rho^2\sqrt{q_0-R^2/\rho^2}}\right),\\\label{aeq004}
q_0&=&\frac{1}{p}E\|\hat{\btheta}\|^2_{\bSigma},\\\label{aeq005}
q&=&\frac{1}{p\sqrt{\xi_0}}E\left\langle\bSigma^{1/2}\bz,\hat{\btheta}\right\rangle\\\label{aeq006}
R&=&\frac{1}{p}E\langle\Sigma\btheta_\star,\hat{\btheta}\rangle.
\end{eqnarray}
Note that two types of Gaussian random variables are introduced, one is in primary $\hat{\btheta}$ and another one is in conjugate $\hat{u}$. The variances of these two random variables are controlled by $\xi_0$ and $q_0$ respectively. It is interesting to see that $\xi_0$ is determined by the expectation over a quadratic form of $\hat{u}$ while $\xi_0$ is determined by the expectation over a quadratic form of $\hat{\btheta}$.

The above formulas are for general positive definite covariance matrix $\bSigma$. Then after applying the random features model and integrating over $\bz$, we obtain the explicit nonlinear equations (\ref{aeq004}), (\ref{aeq005}), and (\ref{aeq006}) for determining six parameters $q_0~,q$, and $R$ as
\begin{eqnarray}\label{beq004}
q_0&=&\frac{1}{p}\xi_0Tr\left(\bSigma^{1/2}(\xi\bSigma+\tau \bI_p)^{-1}\bSigma(\xi\bSigma+\tau \bI_p)^{-1}\bSigma^{1/2}\right)\\\label{beq005}
&&+\frac{1}{p}\hat{R}^2(\btheta_\star)^T\bSigma(\xi\bSigma+\tau \bI_p)^{-1}\bSigma(\xi\bSigma+\tau \bI_p)^{-1}\bSigma\btheta_\star\\\label{beq006}
&=&\xi_0f_2(\xi,\tau)+\hat{R}^2\rho^2f_3(\xi,\tau),\\\nn
R&=&\hat{R}\rho^2f_1(\xi,\tau),\\\nn
q&=&f_0(\xi,\tau),
\end{eqnarray}
where
\begin{eqnarray}\nn
f_0(\xi,\tau)=\int\frac{X}{\xi X+\tau}\mu(dX,dW),&&f_1(\xi,\tau)=\int\frac{W^2X}{\xi X+\tau}\mu(dX,dW),\\\nn
f_2(\xi,\tau)=\int\frac{X^2}{(\xi X+\tau)^2}\mu(dX,dW),&&f_1(\xi,\tau)=\int\frac{W^2X^2}{(\xi X+\tau)^2}\mu(dX,dW).
\end{eqnarray}
After variable substitution $R/\rho\rightarrow R$ and $\rho\hat{R}\rightarrow\hat{R}$,  we derive the equations (\ref{eq001})-(\ref{eq006}) in the main text.  

\section*{Proof of Corollary \ref{coro1}}

Under $\tau=0$, from (\ref{beq004}), (\ref{beq005}), and (\ref{beq006}), we have
\begin{eqnarray}\nn
q_0=\frac{\xi_0+\hat{R}^2\rho^2}{\xi^2},~q=\frac{1}{\xi},~R=\frac{\hat{R}\rho^2}{\xi^2}.
\end{eqnarray}
Substitute into (\ref{aeq001}), (\ref{aeq002}), and (\ref{aeq003}), we have
\begin{eqnarray}\label{p1}
q_0-\frac{R^2}{\rho^2}&=&\alpha\phi_3,\\\label{p2}
1&=&-\frac{\alpha\phi_2}{\sqrt{q_0-R^2/\rho^2}},\\\label{p3}
\frac{R}{\rho}&=&\alpha\left(\phi_1-\frac{R\phi_2}{\rho\sqrt{q_0-R^2/\rho^2}}\right).
\end{eqnarray}
Substituting (\ref{p2}) into (\ref{p3}), we have $\phi_1=0$. From (\ref{p1}), we have
\begin{eqnarray}\nn
q_0-\frac{R^2}{\rho^2}&=&\alpha E\left\{\left(\hat{u}-y^\star u^\star R/\rho-\sqrt{q_{0}-R^2/\rho^2}y^\star z\right)\left(\hat{u}-y^\star u^\star R/\rho-\sqrt{q_{0}-R^2/\rho^2}y^\star z\right)\right\},
\end{eqnarray}
where $u^\star\perp z,~u^\star\sim N(0,1),~z\sim N(0,1)$, and $P(y=+1|u^\star)=f_+(\rho u^\star)$. Substituting (\ref{p2}) and (\ref{p3}), we obtain 
\begin{eqnarray}\nn
E\left\{\left(\hat{u}-y^\star u^\star R/\rho-\sqrt{q_{0}-R^2/\rho^2}y^\star z\right)\hat{u}\right\}=0.
\end{eqnarray}
Denote $r=R/\rho/\sqrt{q_0}$. For SVM, we get
\begin{eqnarray}\nn
0&=&E\left\{\left(1-\sqrt{q_{0}}(ry^\star u^\star+\sqrt{1-r^2}y^\star z)\right)I(1-q\le\sqrt{q_{0}}(ry^\star u^\star+\sqrt{1-r^2}y^\star z)\le 1)\right\}\\\nn
&&+E\left\{q\left(q+\sqrt{q_{0}}(ry^\star u^\star+\sqrt{1-r^2}y^\star z)\right)I(\sqrt{q_{0}}(ry^\star u^\star+\sqrt{1-r^2}y^\star z)\le 1-q)\right\}.
\end{eqnarray}
We are interested in the separability, i.e. the behaviour of $q_0\rightarrow\infty$. The above equation implies that $q/\sqrt{q_0}\rightarrow\infty$. Therefore from (\ref{p1}) and (\ref{p3}), we obtain
\begin{eqnarray}\label{s1}
1/\alpha&=&E\left\{\left(\frac{r}{\sqrt{1-r^2}}y^\star u^\star+y^\star z\right)_+^2\right\},\\\label{s2}
0&=&E\left\{\left(\frac{r}{\sqrt{1-r^2}}y^\star u^\star+y^\star z\right)_+y^\star u^\star\right\},
\end{eqnarray}
which is equivalent to find
\begin{eqnarray}\nn
1/\alpha&=&\min_{c\in\mR}E\left\{\left(cy^\star u^\star+z\right)_+^2\right\}.
\end{eqnarray}

\section*{Proof of Proposition \ref{prop2}}
From equations (14), (15), and (16) in Proposition 3 of \cite{huang2019large}, we obtain
\begin{eqnarray}\nn
q_0-\frac{R^2}{\gamma^2}&=&\alpha E\{(\hat{u}-a)^2\},\\\nn
\frac{R}{\gamma^2}&=&\alpha\mu E(\hat{u}-a),\\\nn
1&=&-\frac{\alpha}{\sqrt{q_0}}E\{(\hat{u}-a)z\},
\end{eqnarray}
where $a=R\mu+\sqrt{q_0}z$. For SVM, define $\gamma^2=\hat{\bmu}^T\bSigma^{-1}\hat{\bmu}$, $z_c=(1-R\mu)/\sqrt{q_0}$, $x=q/\sqrt{q_0}$, and $r=R/\sqrt{q_0}$, we have
\begin{eqnarray}\label{m1}
1-\frac{r^2}{\gamma^2}&=&\alpha\left\{\int_{z_c-x}^{z_c}(z_c-z)^2Dz+x^2\int_{-\infty}^{z_c-x}Dz\right\}\\\label{m2}
\frac{r}{\gamma^2}&=&\alpha\mu\left\{\int_{z_c-x}^{z_c}(z_c-z)Dz+x\int_{-\infty}^{z_c-x}Dz\right\}\\\label{m3}
1&=&\alpha\int_{z_c-x}^{z_c}Dz.
\end{eqnarray}
From (\ref{m1}) and (\ref{m2}), we have
\begin{eqnarray}\nn
1&=&\alpha\left\{\int_{z_c-x}^{z_c}(z_c-z)^2Dz+x^2\int_{-\infty}^{z_c-x}Dz\right\}+\left\{\alpha\gamma\mu\left(\int_{z_c-x}^{z_c}(z_c-z)Dz+x\int_{-\infty}^{z_c-x}Dz)\right)\right\}^2.
\end{eqnarray}
For fixed $\alpha$, $\mu$ has upper bound in order to have solution. Because of (\ref{m3}), the biggest value for $\mu$ we can achieve is when $x\rightarrow\infty$. Therefore the phase transition for Gaussian mixture model is determined by
\begin{eqnarray}\nn
1&=&\alpha\int_{-\infty}^{z_c}(z_c-x)^2Dz+\left\{\alpha\gamma\mu\int_{-\infty}^{z_c}(z_c-z)Dz\right\}^2,
\end{eqnarray}
where $\Phi(z_c)=1/\alpha$.

\bibliographystyle{chicago} 
\bibliography{biblist}
\end{document}